# The classification for High-dimension low-sample size data

Author: Liran Shen[1], Meng Joo Er[1], Qingbo Yin[2, *]

[1] College of Marine Electrical Engineering, Dalian Maritime University, Dalian, 116023, China

[2] College of Information Science and Technology, Dalian Maritime University, Dalian, 116023, China

*Correspondence and requests for materials should be addressed to Qingbo Yin (Email: qingbo@dlmu.edu.cn)

# Abstract (no more than 250 words)

Huge amount of applications in various fields, such as gene expression analysis or computer vision, undergo data sets with high-dimensional low-sample-size (HDLSS), which has putted forward great challenges for standard statistical and modern machine learning methods. In this paper, we propose a novel classification criterion on HDLSS, tolerance similarity, which emphasizes the maximization of within-class variance on the premise of class separability. According to this criterion, a novel linear binary classifier is designed, denoted by No-separated Data Maximum Dispersion classifier (NPDMD). The objective of NPDMD is to find a projecting direction w in which all of training samples scatter in as large an interval as possible. NPDMD has several characteristics compared to the state-of-the-art classification methods. First, it works well on HDLSS. Second, it combines the sample statistical information and local structural information (supporting vectors) into the objective function to find the solution of projecting direction in the whole feature spaces. Third, it solves the inverse of high dimensional matrix in low dimensional space. Fourth, it is relatively simple to be implemented based on Quadratic Programming. Fifth, it is robust to the model specification for various real applications. The theoretical properties of NPDMD are deduced. We conduct a series of evaluations on one simulated and six real-world benchmark data sets, including face classification and mRNA classification. NPDMD outperforms those widely used approaches in most cases, or at least obtains comparable results.

# 1. Introduction

With the accumulation of high-dimensional low sample size data (HDLSS) in various fields of real-world applications (such as image processing and computer vision, data mining, bioinformatics and gene expression) (1-4), classification on these data is a critically important task and attracts much attention during the last few decades. Here, HDLSS indicates that the sample size is less than the feature dimension $d$ ($n << d$, high-dimensional low-sample-size: HDLSS). Moreover, the performance of classical statistical methods for classification on HDLSS are seriously degraded (3-6). The HDLSS data has posed significant challenges to standard statistical methods and have rendered many existing classification techniques impractical (7).

For HDLSS, most study adopted the dimension reduction or regulation as preprocessing before the classification was conducted. These works involve the most of modern classifiers, i.e., Naïve Bayes (NB) (5), Logistic Regression (LR)(4), ensemble methods (6), Linear Discriminant Analysis (LDA) (8), Partial Least Square Regression, Mean Difference (MD) (5), neural network (9) and deep learning (10, 11). These methods with dimension reduction are time-consuming(12).

Meanwhile, a few works conduct the classification on HDLSS without dimensionality reduction. Here, our work concentrate on the methods without dimensionality reduction because these methods can be combined with any pre- or post-processing if necessary. When the sample size $n$ is larger than the feature dimension $d$, linear discriminant analysis(LDA) is a popular and successive method for dimension reduction and classification, which maximizes the so-called Fisher criterion (Rayleigh coefficient) and obtain the projection vector $w$, the largest eigenvectors of $(S_w)^{-1}S_b$ where $S_b$ is the between-class scatter matrix and $S_w$ is the within-class scatter matrix. However, LDA is impractical when $S_w$ is singular due to HDLSS. Li (13) proposed the Maximum Margin Criterion (MMC) to improve the formula of linear discriminant analysis and avoid solving the inverse of the low rank between-class scatter matrix for HDLSS. Support Vector Machine (SVM) (14) is a well-known classifier, which maximize the smallest distance between classes. It can be used directly to any data set, no matter what the *n* is larger or less than *d*. In the practical applications for HDLSS, a phenomenon of data overfitting, so-called "data-piling", is often observed for these classifiers (15, 16), especially SVM. Marron and Qiao et al. proposed the distance-weighted methods (DWD, wDWD and DWSVM) to improve the SVM in the HDLSS setting(16-19). These distance-weighted methods maximize harmonic mean between classes with heavily computing consumption due to second-order cone programming (SOCP), which is more computationally demanding than quadratic programming for SVM (20-22). In Ref (23), a new linear binary classifier PGLMC combined the local structure of the hyperplane and the global statistics information of population with lower computational complexity than the distance-weighted methods owing to solve the similar Quadratic Programming formulation as SVM. Although the distance-weighted methods and PGLMC alleviates the data-piling phenomenon, it is still inevitable to suffer from this overfitting issue for HDLSS even. The proofs for this issue will be given in the following section.

In this paper, a novel linear binary classifier (No-separated Data Maximum Dispersion classifier, NPDMD) was proposed to work directly on HDLSS without dimensionality reduction based on a new discriminant or classification criterion, the *tolerance similarity* (TS). Geometrically, on the premise of being able to distinguish two classes of samples, the farther the samples in each class are in the projection space, the better to avoid the data-piling phenomenon. When we look through the formulation of LDA, it can be found that data-piling phenomenon is an inevitable result when we pursue to minimize intra class differences if $n$ is less than $d$ because the projection vector $w$ falls in the null space of $S_w$. However, the principle of tolerance similarity avoids data-piling by maximizing intra class differences. In term of *tolerance similarity,* NPDMD is designed to find the hyperplane, which can separate samples from two classes and maximize data dispersion in as large an interval as possible. NPDMD holds the same computational complexity as in SVM owing to solve the similar Convex Quadratic Programming formulation. The experimental results demonstrate that our method not only addresses the data-piling issue on HDLSS, but also can be applied to general data with arbitrary dimensions (or the imbalanced data). Our method is stable; no matter the samples are balanced or not across two classes.

The rest of this paper is organized as follows. Section 2 briefly introduces the related methods in HDLSS, especially SVM and the methods based on Distance Weighting. Section 3 elaborates the proposed NPDMD. Section 4 demonstrates experimental results. Finally, Section 5 concludes the paper.

## 2. The relative methods and drawbacks

In this section, we will sketch the reason of data-piling for the related methods without dimensional reduction in HDLSS, which involve MMC, SVM, the methods based on Distance Weighting and PGLMC.

For binary classification problems, a data point in sample space is mapped to a class label chosen from $y$, $\phi: x \to y$, where $x \in R^d$ and $y \in \{+1, -1\}$. Here, for the sake of simplicity, all samples have been centered with mean zero. A linear discriminant function can be denoted as $f(x) = (w^T x + b)$, where the coefficient direction vector $w \in R^d$ has unit $L_2$ norm, and $b \in R$ is the intercept term. Data-piling mean majority of the data lying on two hyper-planes or concentrate on two points after the data are projected to a particular direction vector of a classifier [14]. That is to say, for the sample $x_i$, $w^T x_i = 0$, $,i = 1, 2, \cdots, n$. When $n \ll d$, there are $d - n$ nonzero linear independent solutions for $w$. Therefore, data-piling is a unique phenomenon for HDLSS because there is only a zero solution for $w$ when $n \geq d$ providing the samples are linear independent.

MMC is a variant of LDA, and employs the objective function $\max_{w}[w^T(S_b - S_w)w]$ instead of $(S_w)^{-1} S_b$ to bypass the singular of $S_w$ for HDLSS. This method benefits from the use of sample statistical information. But the data-piling is still a plagued problem due to original LDA' idea, which maximize the differences ($S_b$) between classes and minimize the differences $S_w$ within classes to find a projection vector $w$. This rule is perfect when $d \ll n$ because there are enough samples to provide similarity and dissimilarity between or within classes. However, when $n \ll d$, there are lack of samples to learn the intra-class differences if the rule of $\min_{w}[w^T S_w w]$ is adopted because it is almost true that the solution of $w$ is in the null space of $S_w$.

SVM adopted another route, which try to find a hyperplane with maximum gap between classes. The hyperplane can be represented by the supporting vectors and the projecting direction $w$. In spite of excellent performance in most cases, SVM holds an unexpected trend data-piling when $d \to \infty$ and $n$ is fixed (24). When the samples are from independent and identical distribution population $\pi_i$, and $\pi_i(u_i, \Sigma_i)$ is Gaussian, we should obtain $Var(\|x - u_i\|^2) = 2tr(\Sigma_i^2)$, $x$ is the sample in $i$th class. For HDLSS, all training samples became supporting vectors and concentrated on the boundaries for two classes because

$$\alpha_j = \frac{2}{\psi n_1}\{1 + O(1)\}, \ j = 1, \cdots, n_1 \tag{1}$$

$$\alpha_j = \frac{2}{\psi n_2}\{1 + O(1)\}, \ j = n_1 + 1, \cdots, n \tag{2}$$

Where $n_i$ is the number of samples in $i$th class, $n_1 + n_2 = n$. $\psi = \frac{tr(\Sigma_1)}{n_1} + \frac{tr(\Sigma_2)}{n_2} + \Delta$, $\Delta = \|u_1 - u_2\|^2$. Therefore, since SVM merely use the local structural information (supporting vectors) of the training samples without considering the statistical information of training samples, data-piling is also an inevitable result for HDLSS(12). For detailed proofs about this property of SVM on HDLSS, please refer to (24).

The methods based on Distance Weighting were developed in recent years, which maximizes the harmonic mean of the distances of all data vectors. While they alleviate the data-piling phenomenon or overfitting to a certain extent, there are still a few issues (1) computing consumption. The current state of the art implementation for these methods is based on second-order cone programming (SOCP), which is more computationally demanding than quadratic programming(20). (2) Data-piling cannot be completely avoided. For the latter issue, the proof is as follow:

For the original objective function of DWD

$$\operatorname*{argmin}_{w,b} \sum_{i=1}^{n} \left(\frac{1}{r_i} + C\eta_i\right) \tag{3}$$

subject to $r_i = y_i(x_i^T w + b) + \eta_i$, $r_i \geq 0, \eta_i \geq 0$, $\|w\|^2 \leq 1$

For simplicity, we omit the relaxation term $C\eta_i$ when the samples from two classes are completely separable. Assuming that $L = \sum_{i=1}^{n} \left(\frac{1}{r_i}\right), L \geq \varepsilon$. $\exists \varepsilon \in [a,b)$, $a$ and $b$ are the minimum and maximum of positive real set $\left\{\left(\frac{n}{r_i}\right) \geq 0, i = 1, \cdots, n\right\}$. According to the mean value inequality, it can be true that $\varepsilon = n \cdot \sqrt[n]{\prod_{i=1}^{n} \frac{1}{r_i}}$ and $\varepsilon = n\frac{1}{r_c}$ when $r_c = r_1 = r_2 = \cdots = r_n$, i.e. $L = n\frac{1}{r_c}$ when $r_c = r_1 = r_2 = \cdots = r_n$. Here, we adopted the same limit value $r_c$ for positive and negative classes. However, this does not affect the conclusion. Because even if we adopt different limit values $r_{c+}$ and $r_{c-}$ for positive and negative classes, we will still get similar results of data-piling.

The proof is completed. Therefore, DWD is still with the gradual trend of data-piling in HDLSS.

PGLMC attempted to find the projecting direction $w$, on which the distances between the projecting points from two classes are as far as possible and that the gap (minimum distance) between two classes is as large as possible. PGLMC merely uses the item $(u_1 - u_2)$ to control the differences between classes, and does not consider the intra-class differences. Therefore, as the methods based on Distance Weighting do, PGLMC only alleviate the issue of data-piling instead of overcoming it.

According to the above description, it can be known that all of the above methods only emphasize the difference between classes and the similarity within a class, but ignore the difference between samples within a class. In practical applications, when the sample size is enough ($d \ll n$), these methods have excellent performance. In the case of HDLSS, they are biased and not stable because there are plenty of disturbing clues to meet the similarity criteria $\min_{w}[w^T S_w w]$, which only emphasizes the maximization of similarity.

## 3. The proposed method

In the view of consideration of the above-mentioned and other results in the literature (25-27), we propose a new classification criterion on HDLSS, *tolerance similarity,* which involves two rules: (1) Class separability. In theory, there is at least a hyperplane to separate clearly the samples for training to classes. (2) The similarity and difference of samples within a class must be taken into account. That is to say, the similarity of samples within a class has tolerance (difference). It has already known that LDA and MMC leverage $\min_{w}[w^T S_w w]$ to maximize the similarity, which discards the difference of intra-class samples and leads to data-piling on HDLSS because this design has no tolerance for sample difference within classes. In view of this, on the premise of class separability, $\max_w[w^T S_w w]$ is a good choice on HDLSS instead of $\min_w[w^T S_w w]$ to measure the similarity with tolerance difference.

Here, a linear classifier is conceived based on *tolerance similarity* to maximize the dispersion interval of all training samples in the projection space as the gap between two classes is maximized.

### 3.1 No-separated Data Maximum Dispersion Linear Classifier

The objective function of no-separated data maximum dispersion linear classifier (NPDMD) is as follow:

$$\min_w \left( \frac{\|w\|^2}{w^T S_W w} + C_0 \sum_{i=1}^n \xi_i \right) \quad (4)$$

$$\text{s.t.} \, y_i(w^T x_i + b) \geq 1 - \xi_i, i = 1,2,\cdots,n$$

$$\xi_i = \ell(y_i f(x_i)), \, \xi_i \geq 0$$

Where

$$S_W = \Sigma_1 + \Sigma_2 \quad (5)$$

$$\Sigma_j = \frac{1}{n_j} \sum_{x \in j_{class}} (x - u_j)(x - u_j)^T \quad (6)$$

According to *tolerance similarity*, the numerator $\|w\|^2$ is in charge of rule 1 and minimized to separate the samples from two classes by maximizing the gap between two classes. This is derived from SVM (14). The denominator term $w^T S_W w$ is for rule 2 and also maximized to control training samples from two classes as far as possible along the projecting direction $w$, and balance the covariance between two classes. The hinge loss $\ell(\vartheta) = (1-\vartheta)_+$, where $(\vartheta)_+ = \vartheta$ if $\vartheta \geq 0$ and 0 others. $u_j$ is the mean of training samples from $j$th class, $j = 1,2$.

To solve (4), the Lagrangian formulation can be written as

$$L(w,b,\alpha) = \frac{\|w\|^2}{w^T S_W w} + C_0 \xi^T \mathbf{1} + \alpha^T[\mathbf{1} - \xi - Y(Xw + b\mathbf{1})] - \mu^T \xi \tag{7}$$

where $Y$ is the $n \times n$ diagonal matrix with the components of $y_i$ on its diagonal; $X \in R^{n \times d}$, which $i$th row is sample $x_i$; $w \in R^{d \times 1}$ is the projecting vector; $\mathbf{1}$ is the column vector 1; $\alpha = (\alpha_1, \cdots, \alpha_n)^T \in R^{n \times 1}$, $\alpha_i > 0$ and $\alpha_i$s are Lagrangian multipliers; $\mu = (\mu_1, \cdots, \mu_n)^T \in R^{n \times 1}$, $\mu_i > 0$ and $\mu_i$s are Lagrangian multipliers; $\xi = (\xi_1, \cdots, \xi_n)^T \in R^{n \times 1}$.

It has been proven that $\frac{1}{w^T S_W w}$ and $[C - w^T S_W w]$ are with same effect in the optimization formula(12). The formula (4) can be reformulated to facilitate calculation as below:

$$L(w,b,\alpha) = \frac{1}{2}\|w\|^2 + \frac{1}{2}\lambda(C - w^T S_W w) + \alpha^T[\mathbf{1} - \xi - Y(Xw + b\mathbf{1})] + C_0 \xi^T \mathbf{1} - \mu^T \xi \tag{8}$$

By differentiating the Lagrangian formulation with respect to $w$, $b$ and $\xi$, we can get the following conditions:

$$\frac{\partial L}{\partial w} = w - \lambda S_W w - X^T Y^T \alpha = 0$$

$$w = (I - \lambda S_W)^{-1} X^T Y^T \alpha \tag{9}$$

$$\frac{\partial L}{\partial \xi} = C_0 \mathbf{1} - \alpha - \mu, \ C_0 \mathbf{1} = \alpha + \mu \tag{10}$$

$$\frac{\partial L}{\partial b} = \alpha^T Y \mathbf{1} = 0 \tag{11}$$

When substituting (9), (10) and (11) into (8), we can obtain the dual form as follow

$$L(\alpha) = -\frac{1}{2}\alpha^T Y X (I - \lambda S_W)^{-1} X^T Y^T \alpha + \alpha^T \mathbf{1} + \frac{1}{2}\lambda C \quad (12)$$

When provided that

$$H = Y X (I - \lambda S_W)^{-1} X^T Y^T \quad (13)$$

$H$ is a symmetric positive semidefinite matrix, $\tau = \mathbf{1}$ is a column vector with 1, the item $\frac{1}{2}\lambda C$ does not contain variable $\alpha$ and can be discarded, the above equation (12) can be

$$L(\alpha) = -\frac{1}{2}\alpha^T H \alpha + \alpha^T \tau \quad (14)$$

So, the optimization problem (4) can be transformed into the following

$$\underset{\alpha}{\arg\max}\, L(\alpha) \quad (15)$$

$$\text{s.t. } \alpha^T Y \mathbf{1} = 0, C_0 \geq \alpha_i \geq 0$$

The above formula is a classical quadratic programming problem. From the formulation (4), it can be known that the Karush-Kuhn-Tucker condition must be satisfied as follow:

$$\alpha_i \geq 0,\ \mu_i \geq 0$$

$$y_i(w^T x_i + b) - 1 + \xi_i \geq 0$$

$$\alpha_i [y_i(w^T x_i + b) - 1 + \xi_i] = 0$$

$$\xi_i \geq 0,\ \mu_i \xi_i = 0$$

The formulation (15) can be regarded as a quadratic programming problem with equality constrains while inequality conditions are just looked upon as to scale the coefficients $\alpha_i$. The intercept term $b$ can be obtained by the Criterion of Minimum misclassified samples as follows:

$$\underset{b}{\operatorname{argmin}} J(b) = \sum_i^N sgn[-1 * y_i(w^T x_i + b)] \qquad (16)$$

$$sgn(x) = \begin{cases} +1, & x \geq 0 \\ -1, & x < 0 \end{cases} \qquad (17)$$

Given a new sample $x$, the classifier of NPDMD is defined by

$$\hat{y} = f(x) = (w^T x + b) = [(I - \lambda S_W)^{-1} X^T Y^T \alpha]^T x + b \qquad (18)$$

From the above formulation, it can be known that when $\lambda = 0$, NPDMD is identical to SVM. In other words, SVM is a special case of NPDMD. And the concept of support vector is still valid, and defined as $S = \{x_i | y_i(w^T x_i + b) = 1 - \xi_i \text{ and } \alpha_i > 0\}$. The projection vector $w$ is composed of two parts. The first part is about the statistics of sample population $(I - \lambda S_W)$, which is in charge of the preferable shape. The second part is the sample matrix $X^T Y^T$ and the weight vector $\alpha$, which represents the selection of samples or sample sparsity. Moreover, the term $(I - \lambda S_W)$ ensures that the optimization formula (14) can be solved in the whole feature space.

The performance properties of NPDMD is deduced in three different flavors: (1) Fisher consistency, (2) Asymptotic under extremely imbalanced data, (3) High Dimension low sample size asymptotic. Detailed information about the theoretical properties of NPDMD is discussed in the Supplementary material.

**3.2 Accelerated Extension of NPDMD**

For formulation (13), we must calculate the inverse of $d \times d$ matrix $(I - \lambda S_W)$, which is a time consuming problem for HDLSS ($d \gg n$). We should simplify the involved computation. $S_W$ can be decomposed as follow:

$$S_W = \Sigma_1 + \Sigma_2 = B^T K B \tag{19}$$

$$K = diag(\left[\frac{1}{1*n_1}, \cdots, \frac{i}{i*n_1}, \cdots, \frac{n_1}{n_1*n_1}, \frac{n_1+1}{(n_1+1)*n_2}, \cdots, \frac{n_1+n_2}{(n_1+n_2)*n_2}\right])$$

Where $B$ is a $n \times d$ matrix, which $i$th row is the sample $x_i - u_j$. Then, it can be known

$$(I - \lambda S_W) = I - \lambda B^T K B \tag{20}$$

When we resort to the Sherman-Morrison-Woodbury (SMW) identity (28)

$$(A + UCV)^{-1} = A^{-1} - A^{-1}U(C^{-1} + VA^{-1}U)^{-1}VA^{-1} \tag{21}$$

to compute the inverse of $(I - \lambda S_W)$, the following formula is used to transform the original matrix to SMW form.

$$(I - \lambda S_W)^{-1} = [I + B^T(-\lambda K)B]^{-1} \tag{22}$$

Providing $A = I$, $U = B^T$, $C = -\lambda K$ and $V = B$

$$(I - \lambda S_W)^{-1} = [I + B^T(-\lambda K)B]^{-1} = I - I(B^T)[(-\lambda K)^{-1} + BB^T]^{-1}B$$

$$(I - \lambda S_W)^{-1} = I - B^T[(-\lambda K)^{-1} + BB^T]^{-1}B \tag{23}$$

From Equation (23), it can be noticed that both of $K$ and $BB^T$ are the $n \times n$ matrix. Furthermore, $d \times d$ matrix $(I - \lambda S_W)^{-1}$ can be calculated by the inverse of $n \times n$ matrix $[(-\lambda K)^{-1} + BB^T]^{-1}$. For HDLSS ($d \gg n$), the computation cost of $(I - \lambda S_W)^{-1}$ can be notably reduced.

**3.3 Computation Complexity**

The computation complexity for the objective function of NPDMD involve two parts (1) $d \times d$ matrix $(I - \lambda S_W)^{-1}$. (2) Quadratic Programming formulation for Equation (15).

For $d \times d$ positive semidefinite matrix, the $d$ pairs of eigenvalue and eigenvector can be computed in $O(d*d)$ time (29) (30). In the HDLSS case, with $d \gg n$, we adopt the accelerated extension of NPDMD to reduce the computation cost to $O(n*n)$ time.

Convex quadratic programming (QP) can be solved in polynomial time with either the ellipsoid or interior point method. QP's running time is $O(n^{1/2})$ iterations, each iteration requiring $O(n^3)$ arithmetic operations on integers (31).

For SOCP with efficient primal-dual interior point method, it requires is $O(n^{1/2})$ iterations, each requiring $O(n^2 max\{n,d\})$ operations (16, 32). In the HDLSS case, with $d \gg n$, the computation cost for DWD would be greater than in the SVM and NPDMD.

**4. Experimental results**

To evaluate the proposed NPDMD algorithm, in this section, we perform a series of experiments systematically on both simulation data and real-world classification problems. First, we present one synthetic data set for clearly comparing NPDMD with DWD, wDWD SVM and PGLMC. Second, on real-world problems, six data sets depicted in Ref (33-39) are used to evaluate the classification accuracies.

In the following experiments, to eliminate the dependence of the results on the particular training data used, some measures are defined and the average (or mean) of these measures are reported, which are obtained for different randomly sample splits. And the programs were developed in MATLAB and R, and executed in Inter I7-9700 Processor 3.6G Hz system with 64GB RAM. For the methods based on Distance weighting, we adopt the linear binary implementations in R package 'kerndwd'(20).

**4.1 Measures of Performance**

In order to evaluate the performance of NPDMD and compare it fairly with other methods, we adopted general performance measures (confusion matrix, ROC curve) and specific measures for HDLSS (*correct classification rate*: CCR, *mean within-group error*: MWE, and *angle*), which were used in reference (23, 25). In general, the Bayer rule classifier was as the benchmark in HDLSS for comparison. Here, the Bayes direction (25), which serves as the benchmark to be compared with, is only theory direction. In the below simulation setting, we suppose that data are sampled from two multivariate normal distribution with different mean vectors ($\mu_+$ and $\mu_-$) and same covariance matrices $\Sigma$. We can get the following Bayes rule for simulation dataset

$$sign(x^T w_{Bayes} + b_{Bayes}) \tag{23}$$

$$w_{Bayes} = \Sigma^{-1}(\mu_+ - \mu_-)$$

$$b_{Bayes} = -\frac{1}{2}w^T(\mu_+ + \mu_-)$$

Since the theory distribution of samples cannot be known in real applications, Bayes rule is only adopted in simulation experiment to analyze the experimental results. CCR is a measure of classification performance when the balanced data for two classes is involved. MWE is a measure while the imbalanced data are involved. When data is balanced, $CCR = 1 - MWE$. *Angle* is measured by the difference between the estimated discrimination direction $w$ and the Bayes rule direction $w_{Bayes}$: $\angle(w, w_{Bayes})$. *Angle* is generally for interpretability performance. The interpretability is an uncertain concept and measured by the angle between the discriminant direction $w$ of classifier under investigation and of the Bayes classifier (23). Generally speaking, it is reasonable that the closer to the Bayes rule direction, the better the interpretability is (25).

### 4.2 Simulations Data: Experiment 1

A simulated data set is used to compare the classification and interpretability performance among the NPDMD, the original SVM, DWD, wDWD and PGLMC. The classification performances are measured for a large test data set with 3000 observations (1500 for each class). The process is in accordance with the literature (19, 23, 25).

The simulation settings are as below: constant mean difference $\mu \equiv c\mathbf{1}_d$, identity covariance matrix $\Sigma \equiv I_d$, where $c > 0$ is a scaling factor with $2c\|\mathbf{1}_d\|_2 = 2.7$. This setting represents the Mahalanobis distance between the two classes and a reasonable difficulty for classification according to the Bayes rule. In this setting, all samples from two classes are from multivariate normal distributions $N_d(\pm\mu, \Sigma)$.

In training stage, the positive class sample size be 120 and the negative class sample size be 90. The imbalance factor $m \geq 0$ is defined as the sample size ratio between the majority class and minority class. Here, $m = 1.33$. We vary the dimension $d$ in {80, 150, 240, 650, 900, 1500, 2400}, thus last five cases correspond to HDLSS setting.

In Figure 1, we report the comparison results of five replications among DWD, wDWD, SVM, PGLMC and NPDMD. The test data set is with 1500 samples in each class. The boxplots in Figure 1(a) are about the scatter intervals of the correct classification rate for five methods. Figure 1(b) is the mean curve of CCR for five methods. NPDMD is the best one in all dimensions, and the performance of CCR for wDWD, SVM and PGLMC gradually tend to be consistent. While the dimensionality increases, the performance superiority of CCR for NPDMD becomes more and more obvious. Figure 1(c) and (d) are about the mean within-group errors for five methods and their mean curves. It is obvious that according to the measure of MWE, NPDMD is also the best one. Figure 1(e) is about the mean curves of angle differences between the estimated discrimination direction $w$ and the Bayes rule direction for five methods. NPDMD is still optimal, and all methods are gradually becoming consistent while the dimensionality increases.

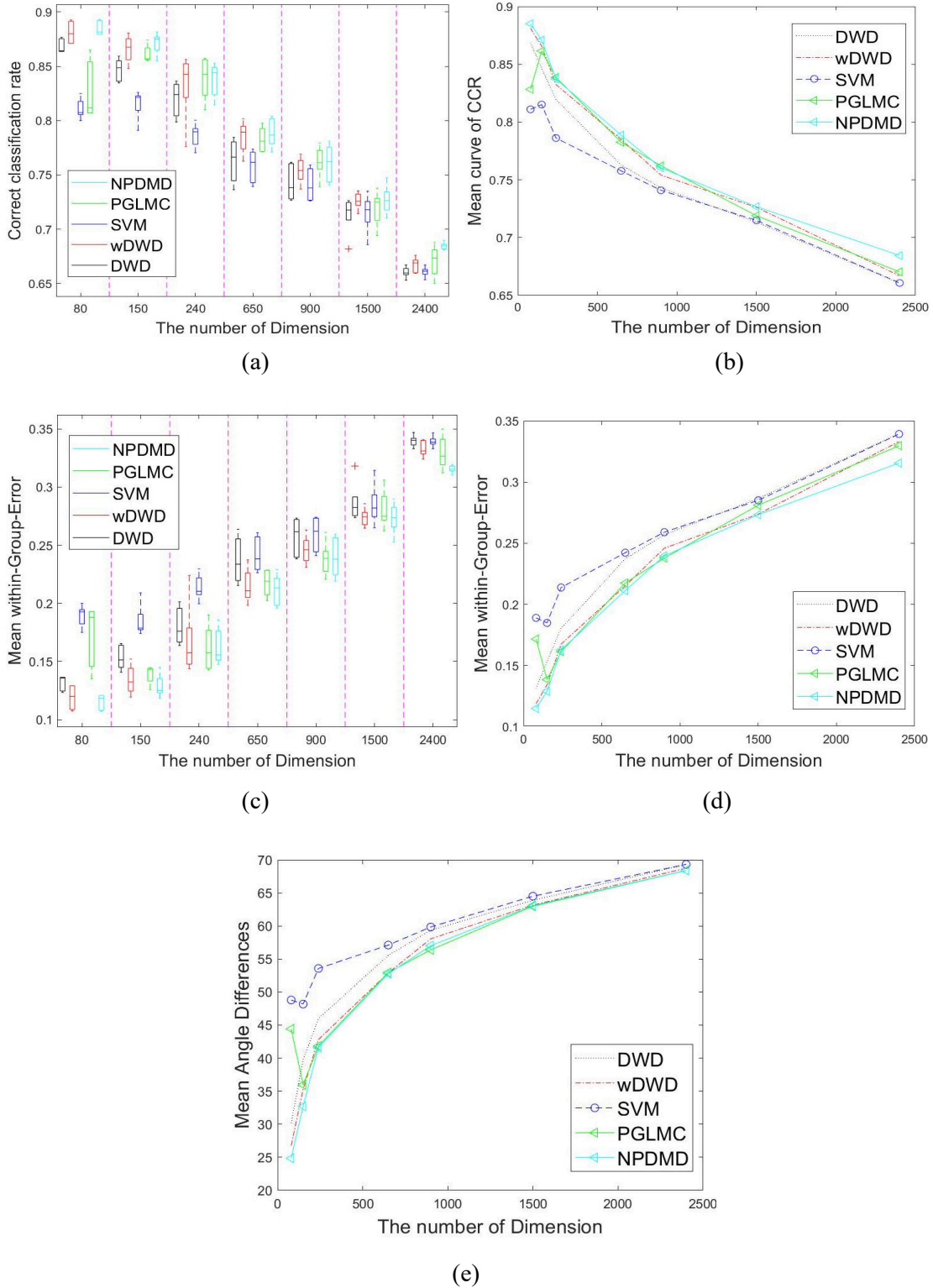

Figure 1. Comparison among five methods for simulation experiment 1 with 5 replications. (a) The boxplots of CCRs. (b) The mean curve of CCRs. (c) The boxplot for the mean within-group error. (d) The mean curve of MWE. (e) The mean curves of angle differences between the estimated discrimination direction and the Bayes rule direction.

In Figure 1, provided that all of these measures are analyzed simultaneously, it can be found that NPDMD gradually obtains the performance superiority of CCR and WME as the dimension increases. The plots of projecting the samples with different dimensions to five different discriminant directions (DWD, wDWD, SVM, PGLMC and NPDMD) are in Supplementary Material (from Figure S1 to Figure S7). NPDMD fits simulation data distribution well and the data projection maintains the Gaussian pattern, which implies that there is some potential to interpret the data and generalize this model to new data in terms of NPDMD direction.

**4.3 Real applications**

In this subsection, we demonstrate the performance of NPDMD which are compared with the competing classifiers on six real data sets. These data sets include a face image data set EYaleB (http://vision.ucsd.edu/~leekc/ExtYaleDatabase/ExtYaleB.html) and five gene expression data sets (Alon data set, Shipp data set, Gordon data set, Chowdary data set and Borovecki data set) (https://github.com/ramhiser/datamicroarray).

The characteristics of the data sets used in the experiments are summarized in Table 1. The data dimensions range from 1024 to 22283. Here, when the number of samples in the data sets are much fewer than the dimensionality of each samples (i.e. $n << d$), the data set is definitely with high dimension. Thus, we report ROC, confusion matrix, mean CCR and MWE as the results. For each data set, we report detailed data preparation, evaluation methods and comparison results. Because there are more classes than 2 in the EYaleB data set, we iterate to select one class as positive and the rest classes as negative in the experiments for binary classification.

Table 1. Characteristics of the Data Sets Used in the Experiments

| Type | Data Set | Dim | Class | Examples | Comments |
|---|---|---|---|---|---|
| Image Data | EYaleB | 1024 | 38 | 2414 | Medium sample size |
| High Dim | Alon | 2000 | 2 | 62 | Small sample size |
| | Shipp | 6817 | 2 | 58 | Small sample size |
| | Gordon | 12533 | 2 | 181 | Small sample size |
| | Chowdary | 22283 | 2 | 104 | Small sample size |
| | Borovecki | 22283 | 2 | 31 | Small sample size |

### 4.3.1 Experiment 2: EYaleB data set

In the task of face recognition, many classes and numerous of image features (high dimension) are always involved. The EYaleB data set (1) consists of 2414 images for 38 individuals. For each individual, there are about 64 near frontal images under different illumination conditions with diverse expressions. All images are cropped and resized to $32 \times 32$ pixels as done in (34). The original gray values in the images are used as the features in this experiment. Therefore, the feature vectors for a sample in each class is with 1024 dimensions.

We split all specimens to eleven folds, in which one-fold is for training and the rest ten folds for testing. Parameters for each method are tuned via 3-fold cross-validation within training data. This process is repeated for 4 times. The CCRs, MWEs and ROC curve are exhibited in Fig. 2.

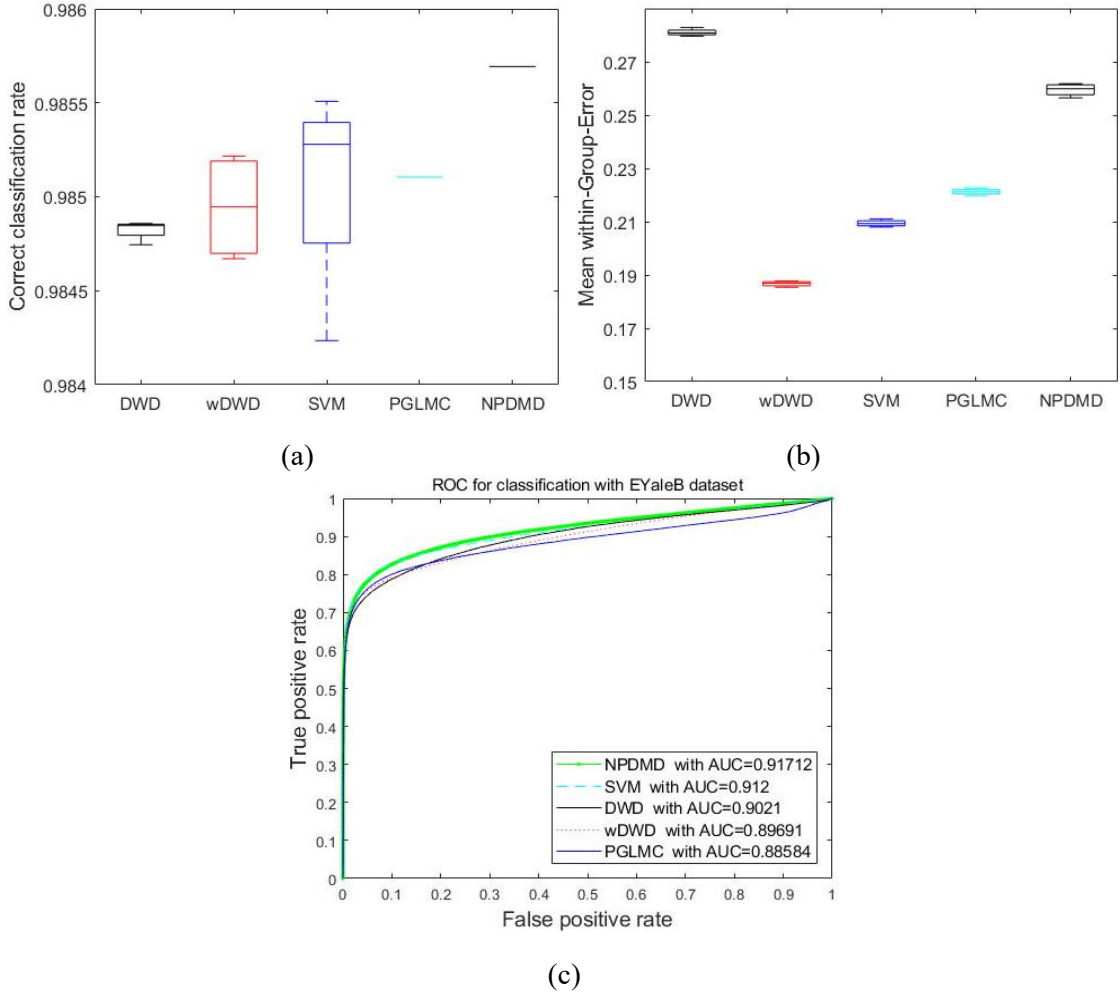

(a)                                    (b)

(c)

Figure 2. Comparison between five methods on EYaleB data. (a) The boxplots of CCRs. (b) The boxplot for the mean within-group error. (c) ROC curves and AUC.

As shown in Fig.2, NPDMD obtains the best CCR and AUC. wDWD obtains the lowest MWE while imbalanced factor $m = 37$ for each individual. The confusion matrix is shown in Table 2.

Table 2. The confusion matrix on EYaleB dataset

| Method | Background | Classification | | CCR (%) | Total CCR (%) | 1-MWE (%) |
|---|---|---|---|---|---|---|
| | | True | False | | | |
| DWD | True | 3570459 | 94125 | 97.43 | 96.08 | 71.75 |
| | False | 53424 | 45638 | 46.07 | | |
| wDWD | True | 3553006 | 19714 | 99.45 | 98.49 | **81.33** |
|  | False | 35528 | 61032 | 63.21 |  |  |
| SVM | True | 3558003 | 14717 | 99.59 | 98.51 | 79.06 |
|  | False | 40047 | 56513 | 58.53 |  |  |
| PGLMC | True | 3559721 | 12999 | 99.64 | 98.49 | 77.87 |
|  | False | 42384 | 54176 | 56.11 |  |  |
| NPDMD | True | 3569401 | 3319 | 99.91 | **98.55** | 74.12 |
|  | False | 49895 | 46665 | 48.33 |  |  |

**4.3.2 Experiment 3: Alon data set**

Alon data set (35) contains gene expression levels of 40 tumour and 22 normal colon tissues for 6500 human genes, which were obtained with an Affymetrix oligonucleotide array. According to Alon setting (35), the 2000 genes with the highest minimal intensity across the samples were reserved for classification.

There are 40 and 22 samples for two classes with imbalance factor $m \approx 1.82$. All specimens were splitted to five folds, in which four-folds are used for training and one fold for testing. Parameters for each method are adjusted via 4-fold cross-validation within training data. This process is repeated for 20 times. The CCRs, MWEs and ROC curve are exhibited in Fig. 3.

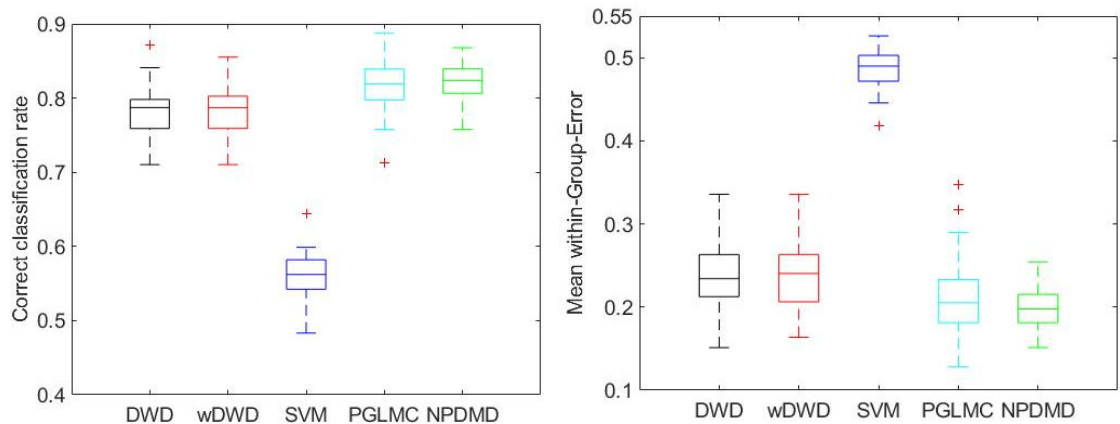

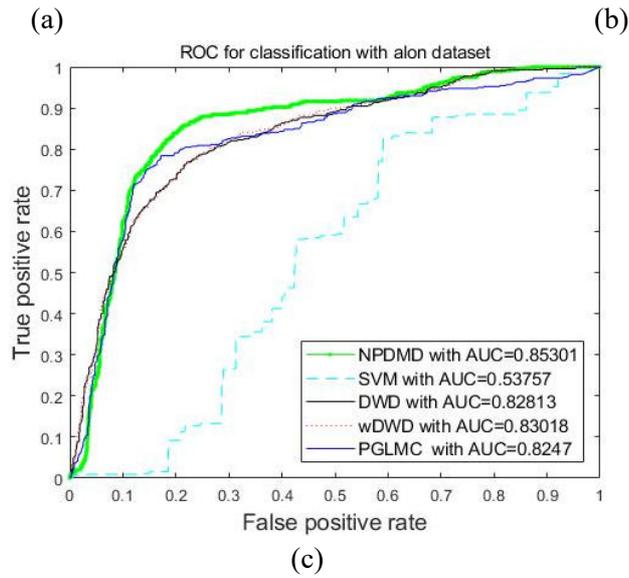

(a)　　　　　　　　　　　　　　(b)

(c)

Figure 3. Comparison between five methods on Alon data. (a) The boxplots of CCRs. (b) The boxplot for the mean within-group error. (c) ROC curves and AUC.

Table 3. The confusion matrix on Alon dataset

| Method | Background | Classification True | Classification False | CCR (%) | Total CCR (%) | 1-MWE (%) |
|---|---|---|---|---|---|---|
| DWD | True | 661 | 147 | 81.81 | 77.73 | 76.07 |
|  | False | 132 | 313 | 70.34 |  |  |
| wDWD | True | 658 | 150 | 81.44 | 77.57 | 76.00 |
|  | False | 131 | 314 | 70.56 |  |  |
| SVM | True | 544 | 256 | 68.00 | 55.97 | 51.05 |
|  | False | 290 | 150 | 34.09 |  |  |
| PGLMC | True | 705 | 95 | 88.12 | 81.37 | 78.61 |
|  | False | 136 | 304 | 69.09 |  |  |
| NPDMD | True | 702 | 98 | 87.75 | **82.26** | **80.01** |
|  | False | 122 | 318 | 72.27 |  |  |

As shown in Figure 3, in this data set, NPDMD is the best one than other methods (DWD, wDWD, SVM and PGLMC) on both CCR and MWE. The confusion matrix is shown in Table 3.

### 4.3.3 Experiment 4: Shipp data set

In Shipp data set (36), there are 58 diffuse large B-cell lymphomas (DLBCLs) patient samples for 6,817 gene expression levels with customized cDNA ('lymphochip') microarrays, which include 32 positive and 26 negative samples with imbalance factor $m \approx 3.05$. All samples were separated to five folds, in which 4-folds are for training and one fold for testing. Parameters for each method are rectified via four-fold cross-validation within training data. We repeated this process for 10 times. The CCRs, ROC and MWEs are exhibited in Fig. 4.

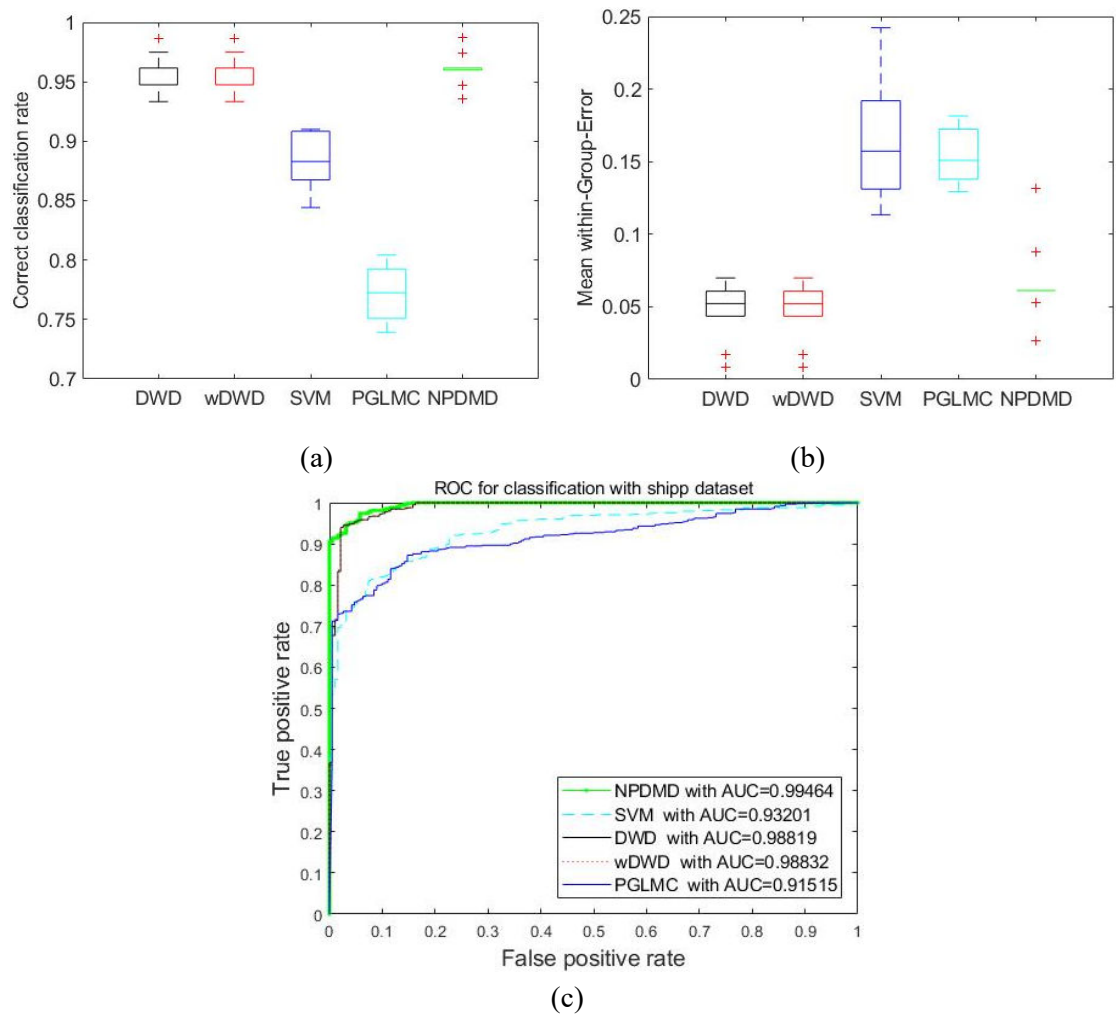

Figure 4. Comparison between five methods on Shipp dataset. (a) The boxplots of CCRs. (b) The boxplot for the mean within-group error. (c) ROC curves and AUC.

As exhibited in Figure 4, NPDMD gets the best CCR and AUC than other methods (DWD, wDWD, PGLMC and SVM), especially superior to SVM and PGLMC. DWD and wDWD obtains the lowest MWE. The corresponding confusion matrix is presented in Table 4.

Table 4. The confusion matrix on Shipp dataset

| Method | Background | Classification True | Classification False | CCR (%) | Total CCR (%) | 1-MWE (%) |
|---|---|---|---|---|---|---|
| DWD | True | 181 | 9 | 95.26 | 95.32 | **95.30** |
|  | False | 27 | 553 | 95.34 |  |  |
| wDWD | True | 181 | 9 | 95.26 | 95.32 | **95.30** |
|  | False | 27 | 553 | 95.34 |  |  |
| SVM | True | 148 | 42 | 77.89 | 86.49 | 83.60 |
|  | False | 62 | 518 | 89.31 |  |  |
| PGLMC | True | 189 | 1 | 99.47 | 77.14 | 84.65 |
|  | False | 175 | 405 | 69.83 |  |  |
| NPDMD | True | 167 | 23 | 87.89 | **96.10** | 93.34 |
|  | False | 7 | 573 | 98.79 |  |  |

### 4.3.4 Experiment 5: Gordon data set

Gordon data set(37) contains 31 tissue samples with malignant pleural mesothelioma (MPM) and 150 tissue samples with adenocarcinoma (ADCA) of Lung microarray study for the expression of 12533 genes, which were obtained assayed using U95A oligonucleotide probe arrays (Affymetrix, Santa Clara, CA).

There are 31 and 150 specimens for two classes with the imbalance factor $m \approx 4.84$. All specimens are splited to five folds, in which 4-folds are for training and one fold for testing. Parameters for each method are adjusted via 4-fold cross-validation within training data. We repeat this process for 16 times. The CCRs, ROC and MWEs are exhibited in Fig. 5.

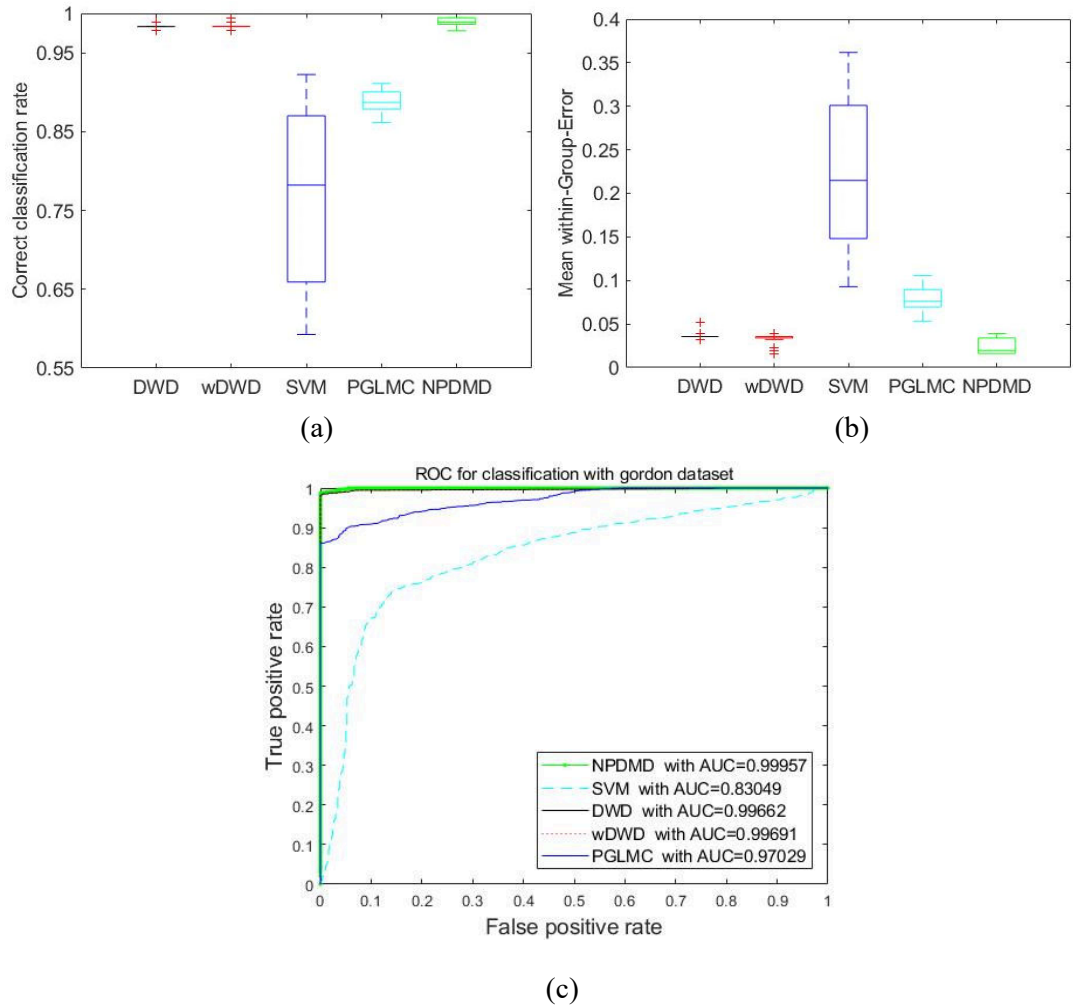

(a)            (b)

(c)

Figure 5. Comparison between five methods on Gordon dataset. (a) The boxplots of CCRs. (b) The boxplot for the mean within-group error. (c) ROC curves and AUC.

As shown in Figure 5, in this data set, NPDMD is the best one than other methods, especially superior to SVM. NPDMD have almost perfect performances. The confusion matrix is in Table 5.

Table 5. The confusion matrix on Gordon dataset

| Method | Background | Classification | CCR (%) | 1-MWE |
|--------|------------|----------------|---------|-------|

|  |  | True | False |  | Total CCR (%) | (%) |
|---|---|---|---|---|---|---|
| DWD | True | 463 | 33 | 93.35 | 98.34 | 96.36 |
|  | False | 15 | 2385 | 99.38 |  |  |
| wDWD | True | 467 | 29 | 94.15 | 98.41 | 96.72 |
|  | False | 17 | 2383 | 99.29 |  |  |
| SVM | True | 397 | 99 | 80.04 | 76.93 | 78.17 |
|  | False | 569 | 1831 | 76.29 |  |  |
| PGLMC | True | 483 | 13 | 97.38 | 88.81 | 92.21 |
|  | False | 311 | 2089 | 87.04 |  |  |
| NPDMD | True | 473 | 23 | 95.36 | **98.93** | **97.51** |
|  | False | 8 | 2392 | 99.67 |  |  |

### 4.3.5 Experiment 6: Chowdary data set

In this experiment, there are 62 tissues with breast tumor and 42 tissues with colon. The samples were assayed using U133A GeneChips (Affymetrix) and data on the expression of 22283 genes (Affymetrix probes) are available (38).

The sample size is 62 and 42 for 2 classes with imbalance factor $m \approx 1.48$. We split all specimens to six folds, in which one-fold is for training and the rest five folds for testing. Parameters for each method are adjusted via 3-fold cross-validation within training data. This process is repeated for 15 times. The CCRs, ROC curve and MWEs are exhibited in Fig. 6.

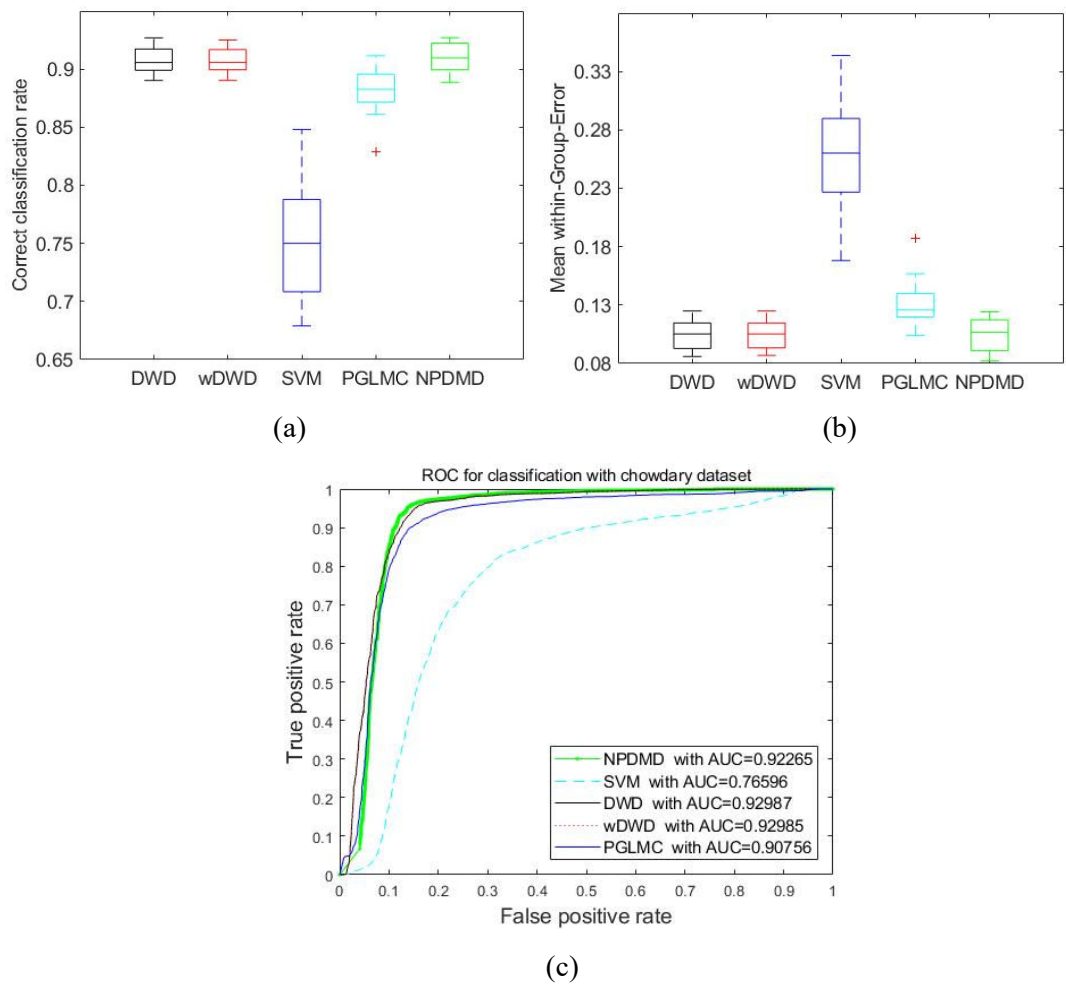

Figure 6. Comparison between five methods on Chowdary dataset. (a) The boxplots of CCRs. (b) The boxplot for the mean within-group error. (c) ROC curves and AUC.

As shown in Figure 6, in this data set, NPDMD is the best one than other methods, especially superior to SVM. The confusion matrix is in Table 6.

Table 6. The confusion matrix on Chowdary dataset

| Method | Background | Classification | | CCR (%) | Total CCR (%) | 1-MWE (%) |
|---|---|---|---|---|---|---|
| | | True | False | | | |
| DWD | True | 2619 | 531 | 83.14 | 90.82 | 89.58 |
| | False | 185 | 4465 | 96.02 | | |
| wDWD | True | 2621 | 529 | 83.21 | 90.81 | 89.58 |
| | False | 188 | 4462 | 95.96 | | |
| SVM | True | 2294 | 856 | 72.83 | 74.73 | 74.42 |
| | False | 1115 | 3535 | 76.02 | | |
| PGLMC | True | 2522 | 628 | 80.06 | 88.21 | 86.89 |
| | False | 292 | 4358 | 93.72 | | |
| NPDMD | True | 2587 | 563 | 82.13 | **91.06** | **89.63** |
| | False | 134 | 4516 | 97.12 | | |

**4.3.6 Experiment 7: Borovecki data set**

This data set contains 17 blood samples with Huntington's disease (HD) and 14 blood samples with matched controls for 22283 genes, which were assayed using U133A GeneChips (Affymetrix) (39).

There are 17 and 14 samples for 2 classes with imbalance factor $m \approx 1.21$. All specimens were splited to five folds, in which 4-folds are for training and one fold for testing. Parameters for each method are tuned via 4-fold cross-validation within training data. We repeated this process for 15 times. The CCRs, ROC curve and MWEs are in Fig. 7.

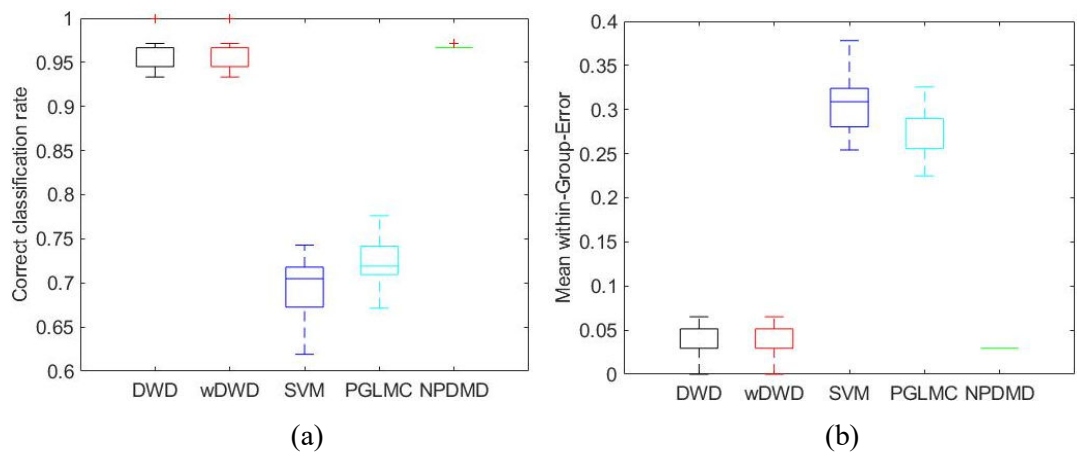

(a)　　　　　　　　　　　　(b)

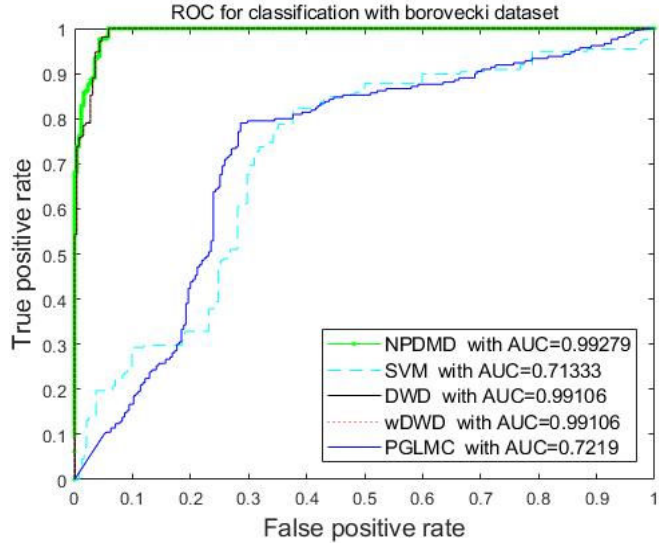

(c)

Figure 7. Comparison between five methods on Borovecki dataset. (a) The boxplots of CCRs. (b) The boxplot for the mean within-group error. (c) ROC curves and AUC.

As shown in Figure 7, in this data set, NPDMD is the best one than other methods, especially superior to SVM. The confusion matrix is in Table 7.

Table 7. The confusion matrix on Borovecki dataset

| Method | Background | Classification | | CCR (%) | Total CCR (%) | 1-MWE (%) |
|---|---|---|---|---|---|---|
| | | True | False | | | |
| DWD | True | 240 | 15 | 94.12 | 96.34 | 96.58 |
| | False | 2 | 208 | 99.05 | | |
| wDWD | True | 240 | 15 | 94.12 | 96.34 | 96.58 |
| | False | 2 | 28 | 99.05 | | |
| SVM | True | 176 | 79 | 69.02 | 69.25 | 69.27 |
| | False | 64 | 146 | 69.52 | | |
| PGLMC | True | 186 | 69 | 72.94 | 72.47 | 72.42 |
| | False | 59 | 151 | 71.90 | | |
| NPDMD | True | 240 | 15 | 94.12 | **96.77** | **97.06** |
| | False | 0 | 210 | 100 | | |

**4.4 Comprehensive analysis**

In this section, we extend the comparisons on the above six data sets with some classical methods such as LR(4, 40), AdaBoost (41, 42), MD(43) and MMC(13, 44), which were ever studied for HDLSS. In order to be able to compare on a fair benchmark, these classical methods also don't consider dimension reduction as pre-processing. For sake of a detailed analysis, we summarize the experimental results in Table 8. We have marked the best results in red. It is demonstrated that NPDMD holds the highest CCR on all of these datasets and best MWE on most of data sets. wDWD gets the best MWE in experiment 2 and 4. Compared with the current methods (DWD, wDWD, SVM and PGLMC), the typical classical methods (LR, AdaBoost, MD and MMC) do not perform well in most cases. It seems that the CCRs of Adaboost and MMC on EYaleB is close or comparable to those of the current methods. However, when we review MWE, it can be found that Adaboost and MMC have little discrimination ability because their MWEs are equal or approximate to 50%, which means that these methods cannot distinguish the samples from two classes. NPDMD shows the best performance on HDLSS. However, if there are many classes in datasets, the performance of NPDMD is disturbed on MWE as experiment 2 except that CCR is still best. It is possible that $w^T S_W w$ is biased because one class is selected as positive and the rest classes as negative in the experiments. Anyway, NPDMD is the best one, and holds more obvious advantage with the higher dimension of data sets.

Table 8. The comprehensive comparison on 6 real datasets

| Experiment | Data | | CCR of Methods (%) | | | | | | | | | 1-MWE (%) | | | | | | | | |
|---|---|---|---|---|---|---|---|---|---|---|---|---|---|---|---|---|---|---|---|---|
| | Dim | Classes | LR | AdaBoost | MD | MMC | DWD | wDWD | SVM | PGLMC | NPDMD | LR | AdaBost | MD | MMC | DWD | wDWD | SVM | PGLMC | NPDMD |
| Experiment 2 | 1024 | 38 | 60.34 | 96.7 | 67.87 | 97.37 | 96.08 | 98.49 | 98.51 | 98.49 | **98.55** | 56.76 | 53.05 | 57.59 | 50.00 | 71.75 | **81.33** | 79.08 | 77.87 | 74.12 |
| Experiment 3 | 2000 | 2 | 61.55 | 64.62 | 69.01 | 51.43 | 77.73 | 77.57 | 55.97 | 81.37 | **82.26** | 60.99 | 50.00 | 71.61 | 48.81 | 76.07 | 76.00 | 51.05 | 78.61 | **80.01** |
| Experiment 4 | 6817 | 2 | 74.97 | 75.33 | 79.42 | 69.88 | 95.32 | 95.32 | 86.49 | 77.14 | **96.10** | 74.61 | 50.00 | 80.14 | 51.25 | **95.30** | **95.30** | 83.60 | 84.65 | 93.34 |
| Experiment 5 | 12533 | 2 | 82.14 | 82.88 | 93.57 | 82.57 | 98.34 | 98.41 | 76.93 | 88.81 | **98.93** | 76.56 | 50.00 | 85.26 | 54.22 | 96.36 | 96.72 | 78.17 | 92.21 | **97.51** |
| Experiment 6 | 22283 | 2 | 82.33 | 59.61 | 71.56 | 63.88 | 90.82 | 90.81 | 74.73 | 88.21 | **91.06** | 81.42 | 50.00 | 65.34 | 56.10 | 89.58 | 89.58 | 74.42 | 86.89 | **89.63** |
| Experiment 7 | 22283 | 2 | 82.06 | 54.76 | 69.59 | 79.87 | 96.34 | 96.34 | 69.24 | 72.47 | **96.77** | 81.11 | 50.00 | 70.06 | 79.11 | 96.58 | 96.58 | 69.27 | 72.42 | **97.06** |

## 5. Conclusion

After analyzing the causes of data-piling HDLSS, we propose a novel classification criterion on HDLSS, tolerance similarity. This criterion stands for the maximization intra-class samples difference rather than the minimization of intra-class samples difference. In the light of this criterion, we proposed a new linear binary classifier NPDMD, which maximize the dispersion interval of all training samples in the projection space when the gap between two classes is maximized. Thanks to this structural design, our method avoids data-piling and exhibits superior performance in HDLSS. From the above experimental results, it can be found that compared with other methods, NPDMD is the best one and holds more obvious advantage with the higher dimension of data sets. The theoretical properties of NPDMD are deduced, which accounts for the performance in the experiments.

The major advantages of this study were four-fold. First, it works well on HDLSS. Second, it combines the sample statistical information and local structural information (supporting vectors) into the objective function to find the solution of projecting direction in the whole feature spaces. Third, it solves the inverse of high dimensional matrix in low dimensional space. Fourth, it is relatively simple to be implemented based on Quadratic Programming. Fifth, it is robust to the model specification for various real applications. The experiment results demonstrated the superiority of NPDMD. And then, NPDMD is with great potential in many applications regardless of HDLSS.

# References

1. Q. Cheng, H. Zhou, J. Cheng, H. Li, A Minimax Framework for Classification with Applications to Images and High Dimensional Data. *IEEE Transactions on Pattern Analysis & Machine Intelligence* **36**, 2117-2130 (2014).
2. Qingbo Yin *et al.*, Integrative radiomics expression predicts molecular subtypes of primary clear cell renal cell carcinoma *Clin Radiol* **73**, 782-791 (2018).
3. L.-F. Chen, H.-Y. M. Liao, M.-T. Ko, J.-C. Lin, G.-J. Yu, A new LDA-based face recognition system which can solve the small sample size problem. *Pattern Recogn* **33**, 1713-1726 (2000).
4. F. S. Kurnaz, I. Hoffmann, P. Filzmoser, Robust and sparse estimation methods for high-dimensional linear and logistic regression. *Chemometr Intell Lab* **172**, 211-222 (2018).
5. A. Bolivar-Cime, J. S. Marron, Comparison of binary discrimination methods for high dimension low sample size data. *J Multivariate Anal* **115**, 108-121 (2013).
6. L. Rokach, Ensemble-based classifiers. *Artificial Intelligence Review* **33**, 1-39 (2010).
7. J. IM, T. DM, Statistical challenges of high-dimensional data. *Philosophical Transactions: Mathematical, Physical and Engineering Sciences* **367**, 4237-4253 (2009).
8. A. Sharma, K. K. Paliwal, Linear discriminant analysis for the small sample size problem: an overview. *International Journal of Machine Learning and Cybernetics* **6**, 443-454 (2015).
9. S. Abpeikar, M. Ghatee, G. L. Foresti, C. Micheloni, Adaptive neural tree exploiting expert nodes to classify high-dimensional data. *Neural Networks* **124**, 20-38 (2020).
10. G. E. Hinton, R. R. Salakhutdinov, Reducing the Dimensionality of Data with Neural Networks. *Science* **313**, 504-507 (2006).
11. Y. LeCun, Y. Bengio, G. Hinton, Deep learning. *Nature* **521**, 436-444 (2015).

12. L. Shen, Q. Yin, Data maximum dispersion classifier in projection space for high-dimension low-sample-size problems. *Knowl-Based Syst* https://doi.org/10.1016/j.knosys.2019.105420, 105420 (2020).
13. H. Li, T. Jiang, K. Zhang, Efficient and robust feature extraction by maximum margin criterion. *Ieee T Neural Networ* **17**, 157-165 (2006).
14. C. Cortes, V. Vapnik, Support-vector networks. *Mach Learn* **20**, 273-297 (1995).
15. J. Ahn, J. S. Marron, The maximal data piling direction for discrimination. *Biometrika* **97**, 254-259 (2010).
16. J. S. Marron, M. J. Todd, J. Ahn, Distance-weighted discrimination. *J Am Stat Assoc* **102**, 1267-1271 (2007).
17. B. X. Wang, H. Zou, Sparse Distance Weighted Discrimination. *J Comput Graph Stat* **25**, 826-838 (2016).
18. X. Y. Qiao, H. H. Zhang, Y. F. Liu, M. J. Todd, J. S. Marron, Weighted Distance Weighted Discrimination and Its Asymptotic Properties. *J Am Stat Assoc* **105**, 401-414 (2010).
19. X. Y. Qiao, L. S. Zhang, Distance-weighted Support Vector Machine. *Stat Interface* **8**, 331-345 (2015).
20. B. X. Wang, H. Zou, Another look at distance-weighted discrimination. *J Roy Stat Soc B* **80**, 177-198 (2018).
21. J. Friedman, T. Hastie, R. Tibshirani, Regularization Paths for Generalized Linear Models via Coordinate Descent. *J Stat Softw* **33**, 1-22 (2010).
22. R. H. Tutuncu, K. C. Toh, M. J. Todd, Solving semidefinite-quadratic-linear programs using SDPT3. *Math Program* **95**, 189-217 (2003).
23. Q. Yin, E. Adeli, L. Shen, D. Shen, Population-guided large margin classifier for high-dimension low-sample-size problems. *Pattern Recogn* **97**, 107030 (2020).
24. Y. Nakayama, K. Yata, M. Aoshima, Support vector machine and its bias correction in high-dimension, low-sample-size settings. *Journal of Statistical Planning & Inference* **191** (2017).
25. X. Y. Qiao, L. S. Zhang, Flexible High-Dimensional Classification Machines and Their Asymptotic Properties. *Journal of Machine Learning Research* **16**, 1547-1572 (2015).
26. P. Hall, Y. Pittelkow, M. Ghosh, Theoretical measures of relative performance of classifiers for high dimensional data with small sample sizes. *J Roy Stat Soc B* **70**, 159-173 (2008).
27. P. Hall, J. S. Marron, A. Neeman, Geometric Representation of High Dimension, Low Sample Size Data. *Journal of the Royal Statistical Society* **67**, 427-444 (2005).
28. G. H. Golub, C. F. V. Loan, *Matrix Computations* (Johns Hopkins University Press, ed. 3rd ed, 1996).
29. I. S. D. a. B. N. Parlett, Orthogonal Eigenvectors and Relative Gaps. *SIAM Journal on Matrix Analysis and Applications* **25**, 858-899 (2003).
30. I. S. Dhillon, B. N. Parlett, Multiple representations to compute orthogonal eigenvectors of symmetric tridiagonal matrices. *Linear Algebra and its Applications* **387**, 1-28 (2004).
31. S. A. Vavasis, "Complexity theory: quadratic programming" in Encyclopedia of Optimization, C. A. Floudas, P. M. Pardalos, Eds. (Springer US, Boston, MA, 2001), 10.1007/0-306-48332-7_65, pp. 304-307.
32. F. Alizadeh, D. Goldfarb, Second-order cone programming. *Math Program* **95**, 3-51 (2003).
33. Q. Yin *et al.*, Associations between Tumor Vascularity, Vascular Endothelial Growth Factor

Expression and PET/MRI Radiomic Signatures in Primary Clear-Cell–Renal-Cell-Carcinoma: Proof-of-Concept Study. *Sci Rep-Uk* **7**, 43356 (2017).
34. A. S. Georghiades, P. N. Belhumeur, D. J. Kriegman, From Few to Many: Illumination Cone Models for Face Recognition under Variable Lighting and Pose. *IEEE Transactions on Pattern Analysis & Machine Intelligence* **23**, 643-660 (2001).
35. U. Alon, . *et al.*, Broad patterns of gene expression revealed by clustering analysis of tumor and normal colon tissues probed by oligonucleotide arrays. *Proceedings of the National Academy of Sciences of the United States of America* **96**, 6745-6750 (1999).
36. M. A. Shipp *et al.*, Diffuse large B-cell lymphoma outcome prediction by gene-expression profiling and supervised machine learning. *Nat Med* **8**, 68-74 (2002).
37. G. J. Gordon *et al.*, Translation of microarray data into clinically relevant cancer diagnostic tests using gene expression ratios in lung cancer and mesothelioma. *Cancer Res* **62**, 4963-4967 (2002).
38. D. Chowdary, J. Lathrop, J. Skelton, Prognostic gene expression signatures can be measured in tissues collected in RNAlater preservative. *Journal of Molecular Diagnostics* **8**, 31-39 (2006).
39. F. Borovecki, . *et al.*, Genome-wide expression profiling of human blood reveals biomarkers for Huntington's disease. *Proceedings of the National Academy of Sciences of the United States of America* **102**, 11023-11028 (2005).
40. E. Makalic, D. F. Schmidt (2011) Review of Modern Logistic Regression Methods with Application to Small and Medium Sample Size Problems. (Springer Berlin Heidelberg, Berlin, Heidelberg), pp 213-222.
41. R. Blagus, L. Lusa, Boosting for high-dimensional two-class prediction. *Bmc Bioinformatics* **16**, 300 (2015).
42. H. Seibold, C. Bernau, A.-L. Boulesteix, R. De Bin, On the choice and influence of the number of boosting steps for high-dimensional linear Cox-models. *Computation Stat* **33**, 1195-1215 (2018).
43. L. Zhang, X. Lin, Some considerations of classification for high dimension low-sample size data. *Statistical Methods in Medical Research* **22**, 537 (2013).
44. Q. Wang *et al.*, Adaptive maximum margin analysis for image recognition. *Pattern Recogn* **61**, 339-347 (2017).

# Figure Legends

Figure 1. Comparison among five methods for simulation experiment 1 with 5 replications.

Figure 2. Comparison between five methods on EYaleB data.

Figure 3. Comparison between five methods on Alon data.

Figure 4. Comparison between five methods on Shipp dataset.

Figure 5. Comparison between five methods on Gordon dataset.

Figure 6. Comparison between five methods on Chowdary dataset.

Figure 7. Comparison between five methods on Borovecki dataset.